\documentclass[conference]{IEEEtran}
\IEEEoverridecommandlockouts
\usepackage{cite}
\usepackage{amsmath,amssymb,amsfonts}
\usepackage{algorithmic}
\usepackage{graphicx}
\usepackage{textcomp}
\usepackage{xcolor}
\usepackage{soul}
\usepackage{comment}
\usepackage{multirow}
\usepackage{booktabs}
\def\BibTeX{{\rm B\kern-.05em{\sc i\kern-.025em b}\kern-.08em
    T\kern-.1667em\lower.7ex\hbox{E}\kern-.125emX}}
\begin{document}

\title{Analysis of voxel-based 3D object detection methods efficiency for real-time embedded systems
}

\author{\IEEEauthorblockN{Illia Oleksiienko and Alexandros Iosifidis}
\IEEEauthorblockA{\textit{Department of Electrical and Computer Engineering, Aarhus University, Denmark} \\
\{io,ai\}@ece.au.dk}
}

\maketitle

\begin{abstract}
Real-time detection of objects in the 3D scene is one of the tasks an autonomous agent needs to perform for understanding its surroundings. While recent Deep Learning-based solutions achieve satisfactory performance, their high computational cost renders their application in real-life settings in which computations need to be performed on embedded platforms intractable. In this paper, we analyze the efficiency of two popular voxel-based 3D object detection methods providing a good compromise between high performance and speed based on two aspects, their ability to detect objects located at large distances from the agent and their ability to operate in real time on embedded platforms equipped with high-performance GPUs. Our experiments show that these methods mostly fail to detect distant small objects due to the sparsity of the input point clouds at large distances. Moreover, models trained on near objects achieve similar or better performance compared to those trained on all objects in the scene. This means that the models learn object appearance representations mostly from near objects. Our findings suggest that a considerable part of the computations of existing methods is focused on locations of the scene that do not contribute with successful detection. This means that the methods can achieve a speed-up of $40$-$60\%$ by restricting operation to near objects while not sacrificing much in performance.
\end{abstract}


\section{Introduction}\label{sec:introduction}
3D object detection is an important task for Autonomous Systems and Robotics, as it provides to the (robotic) agent information needed to perceive its surroundings. Detection of objects needs to be performed in a highly-reliable and real-time manner in order to allow the agent to naturally interact with the environment and avoid collisions. It can be used as the first processing step for path planning, navigation, and/or interaction with objects in the 3D scene. Many methods have been proposed to approach the 3D object detection task, which can be categorized based on the type of data they receive as input: those using monocular images with depth estimation \cite{2018mf3d, 2018monogrnet} or 2D-to-3D regression \cite{2019mono3dobjdet, 2019monodis}, those using binocular images which can provide relative depth of the objects in the scene information \cite{2018pseudolidar, 2019pseudolidar, 2020cdn}, those using point clouds commonly generated by a Lidar sensor \cite{2017voxelnet, lang2018pointpillars, liu2019tanet, 2019hotspotnet}, or hybrid methods combining point clouds with images \cite{2017frustumpointnet, 2019frustumconvnet, 2019pointpainting}. 

Lidar is the most expensive sensor used for 3D object detection, but point cloud-based methods are those providing the best compromise between performance and speed. Point clouds are obtained from firing a set of laser beams and receiving their reflections to calculate exact 3D coordinates of the contact points. 
The generated point cloud is unordered and sparse and, therefore, it cannot be directly processed by regular Convolutional Neural Networks (CNNs) which are the de-facto choice in 2D object detection methods operating on (grid-structured) images. To address this issue, several approaches were proposed to transform the point cloud into a grid-structured format, that can be used as input to CNNs. Projection-based methods use plane projections \cite{2020gslamtc, 2015mvcnn} to create multi-view images of the scene, spherical \cite{2017squeezeseg} or cylindrical \cite{2016velofcn} projections to create a 2D map where each pixel corresponds to a point in a scene. Voxel-based methods select a sub-scene to process and split it into a 3D grid of voxels (volumetric elements) \cite{2017voxelnet} to apply 3D convolutions, or a 2D grid of pillars \cite{lang2018pointpillars, liu2019tanet, 2019hotspotnet} to apply 2D CNNs. While point-cloud based methods are able to achieve good performance in general, class-wise limitations emerge from the increasing sparsity of the point cloud with respect to the distance of the objects from the Lidar sensor, making small objects practically undetectable when they are far away from the Lidar sensor.

In this paper we provide an experimental analysis of the performance of voxel-based methods in relation to the objects' distance from the Lidar sensor. We split the 3D scene used by these methods in two sub-scenes determined by using different depth zones from the Lidar sensor, namely the \textit{near sub-scene} containing the points of objects close to the Lidar sensor (half of the scene along the forward-axis) and the \textit{far sub-scene} containing the points of objects far away from the Lidar sensor (the rest of the scene). We experimentally show that two of the most successful voxel-based methods, i.e. PointPillars \cite{lang2018pointpillars} and TANet \cite{liu2019tanet}, fail to detect small objects appearing in the far sub-scene and that training the models on objects appearing in the near sub-scene leads to performance that is similar or even better than the performance achieved by considering all objects during training. This result indicates that the models trained on all objects in the scene are likely to learn object representations only based on the near objects and try to apply them to objects far away from the Lidar sensor, which are described by a much smaller number of points. Our experimental analysis leads to an important suggestion: in application scenarios involving low-power processing units and requiring real-time operation one should focus on the objects belonging to the near sub-scene, as this leads to a considerable computational cost reduction and the detection rate for small objects belonging to the rest of the scene is low. We observed that following this strategy, a speed-up of $40$-$60\%$ is achieved leading to real-time operation on embedded GPUs.

The remainder of the paper is organized as follows: Section \ref{sec:related-works} provides a description of the related works. Section \ref{sec:methodology} describes the process we follow to define different sub-scenes for 3D object detection based on the respective depth zones and the protocol followed in our experimental analysis. Section \ref{sec:experiments} provides the results of the analysis. Section \ref{sec:conclusions} concludes the paper and formulates directions for future work.

\section{Related work}\label{sec:related-works}
In this Section we provide information regarding the real-time operation of existing 3D object detection methods exploiting Lidar-based point clouds in relation to processing on embedded platforms. Then, we briefly describe the PointPillars \cite{lang2018pointpillars} and the TANet \cite{liu2019tanet} methods which are used in our experimental analysis.

\subsection{Real-time operation in DL-based 3D object detection}\label{sec:real-time-methods}
Deep Learning (DL) based methods gained a lot of attention for solving tasks in Autonomous Systems and Robotics due to their high performance. While for visual-based methods real-time operation is commonly defined at a $30$ FPS inference speed, for Lidar-based methods like those targeting 3D object detection the desired FPS is defined by the specifications of the adopted Lidar sensor. Most available Lidar sensors operate at $10$-$20$ FPS and, thus, Lidar-based methods target an operation at $10$-$20$ FPS as the methods will not be able to process point clouds at a higher speed than it can be generated \cite{2020rtclusterlidar,rashed2019fuseMODNet}. However, even though this choice seems reasonable for isolated application of 3D object detection methods during experiments, Autonomous Systems in real-life applications need to perform a variety of tasks using embedded processing platforms with 3D object detection being a pre-processing step to higher-level analysis tasks, like path planning, navigation and interaction with objects in the scene. Therefore, aiming at the frame rate determined by the Lidar sensor does not lead to satisfactory speed in practice. The availability of high-power embedded platforms like the NVIDIA Jetson AGX Xavier with a powerful GPU and shared CPU-GPU memory (making it a suitable choice for running DL models) allows the adoption of DL-based methods in real-life applications. However, even though such embedded platforms contain powerful GPUs, their capabilities still lack compared to the high-end (desktop) GPUs which are used to develop and test 3D object detection methods. Therefore, efficient method design and usage are needed for the adoption of the DL models in real-life applications involving 3D object detection. Recently, the method in \cite{2020vislidar} proposed a DL model for 3D object detection that operates at $10$ FPS on the NVIDIA Jetson AGX Xavier, which still is far from the commonly considered real-time operation of $30$ FPS.  

3D object detection and tracking methods are frequently based on ideas coming from the much more mature 2D object detection problem. This is due to that many 3D object detection methods use 2D CNNs as backbone networks and, therefore, optimization strategies that target object detection in 2D can be extended to the 3D case too. Speed up approaches that have been proposed for 2D object detection include the use of knowledge distillation to train high-performing compact backbone networks \cite{2020mobilecenternet}, layer pruning to reduce the number of computations in a high-performing backbone network while not sacrificing much in performance \cite{2020efyolo}, and network quantization in which the backbone network is changed by replacing 32/64-bit floating-point operations with faster low-bit integer operations \cite{2020quadaprox, 2020quanttiling}. In this paper we follow a different approach, which focuses on the input data received by Lidar-based methods. The speed-up approaches described above focusing on the efficiency of the backbone networks can be combined with our approach to further increase processing speed, as they are focusing on complementary aspects of the overall system.

\subsection{PointPillars and TANet methods}\label{sec:baselines}
PointPillars \cite{lang2018pointpillars} is one of the fastest Lidar-based 3D Object Detection methods, and it is commonly used as part of other methods \cite{2019hotspotnet, 2019pointpainting, liu2019tanet}. It selects a part of the scene\footnote{Selection of the sub-scene depends on the class of interest and the adopted dataset. KITTI \cite{2012kitti}, which is the most widely used dataset to evaluate 3D object detection methods, provides annotations only for the objects laying inside the part of the scene inside the field-of-view of a camera placed close to the Lidar sensor. Thus, only the frontal part of the point cloud is processed. NuScenes \cite{nuScenes2019}, which is another widely used dataset, has 6 cameras alongside a Lidar, so in every direction there is an input from both cameras and Lidar, allowing to use all points generated by the Lidar. In this paper we follow the setup used in KITTI. Extension of our approach to the setup of NuScenes is straightforward.} in a cuboid shape with boundaries of $([0, x], [-y, y], [z_0, z_1])$, as illustrated in Figure \ref{fig:near-far-example}. In order to transform the (unstructured) point cloud into a grid structure it performs quantization based on a 2D grid along the $x$ (forward-axis) and $y$ (left-right-axis) dimensions to form the so-called pillars. A pillar is a voxel of size $(v_x, v_y, v_z)$ with its size on the vertical-axis being equal to all the available space, i.e. $v_z = z_1 - z_0$. The $x$ and $y$ axes are usually quantized using same-sized  bins, i.e. $v_x = v_y$. Points in pillars are processed to create pillar-wise features that are stored in a pseudo-image where each cell represents a pillar. This image is processed by a Fully Convolutional Network (FCN) \cite{2014fcn} with final classification and regression branches.

TANet \cite{liu2019tanet} is a slower but more accurate method and is a one of the most accurate methods for objects of small size. TANet follows similar processing steps with PointPillars, but it uses a Triple Attention mechanism to create more robust and accurate features for each pillar by combining point-wise, channel-wise and voxel-wise attentions. These pillar features are stored in a pseudo-image in the same way as in PointPillars, but they are processed by a more complex DL model performing Coarse-to-Fine Regression. This DL model consists of a Coarse network with an architecture similar to the Fully-Convolutional Network (FCN) in PointPillars, and a Refine network which uses features from the Coarse network to make more accurate predictions.

The size of a pseudo-image created by both methods depends on the number of pillars that can fit into the scene. For the sub-scene with limits $[0, x]$ along forward-axis and $[-y, y]$ along left-right-axis, the size of the pseudo-image is given by:
\begin{equation}
W = \frac{2y}{v_y} \:\:\:\:\:\:\:\: \textrm{and} \:\:\:\:\:\:\:\: H = \frac{x}{v_x}. \label{eq:pseudo-image-size}
\end{equation}
Increasing the size of pillars, when processing the same scene, leads to a smaller pseudo-image and faster inference. The same effect is obtained by decreasing the size of the sub-scene for fixed-sized pillars. PointPillars and TANet are using FCNs to process the pseudo-image and, therefore, the trained model can be directly applied to pseudo-images of different sizes. However, there is a compromise between fast inference and performance which needs to be considered when selecting the size of the pillars and the size of the scene.

\section{Methodology}\label{sec:methodology}
\begin{figure}[!t]
    \includegraphics[width=1\linewidth]{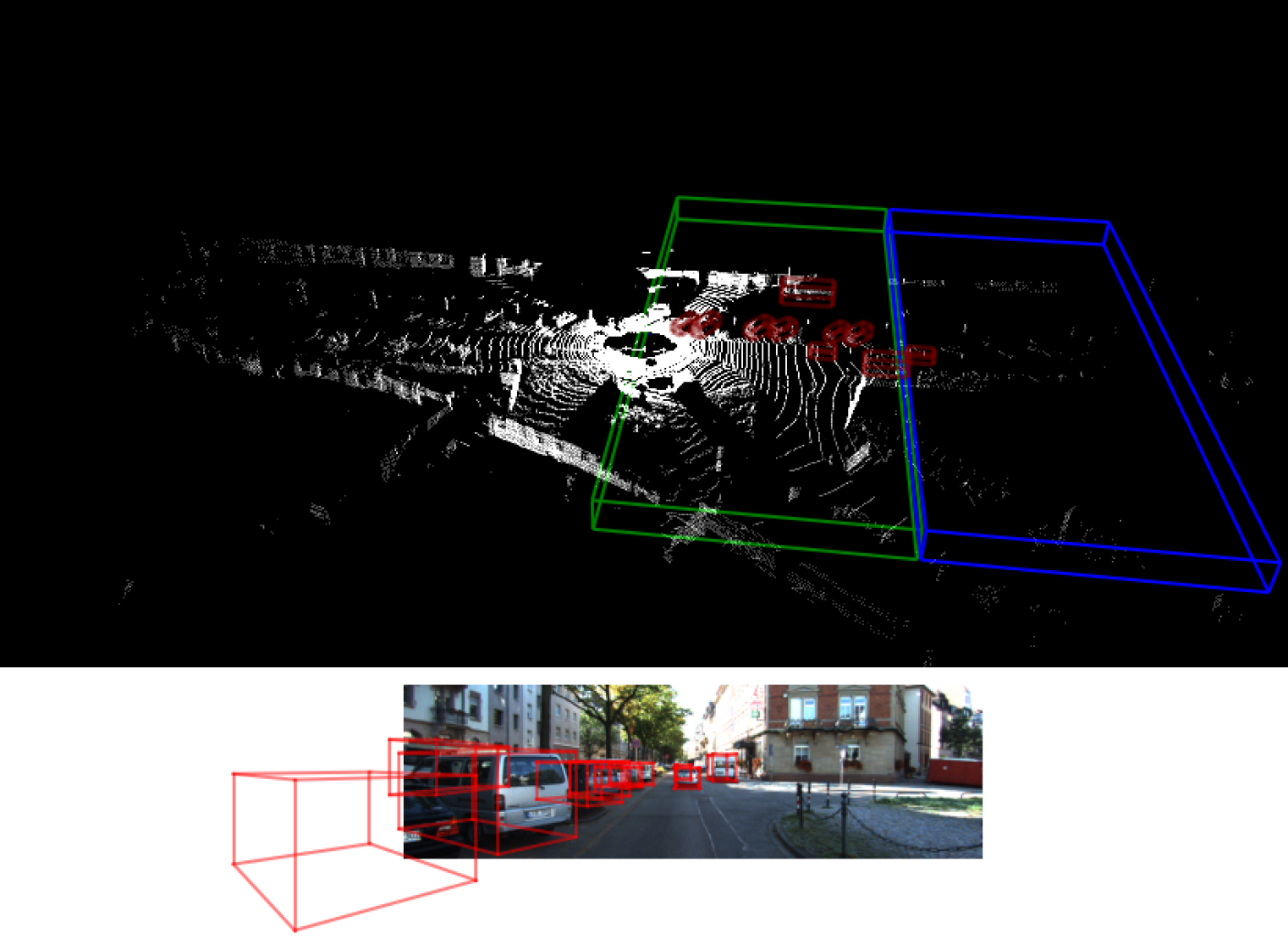}
    \caption{Example of a KITTI frame with a point cloud at the top and the corresponding camera image at the bottom. The Lidar sensor is located at the center of the point cloud. The part of the scene used in PointPillars and TANet (called full-scene in this paper) is the cuboid with boundaries $([0, x], [-y, y], [z_0, z_1])$, which corresponds to the union of the two areas included in the green and blue boxes. We divide this scene into two equally-sized sub-scenes, namely the near sub-scene (with boundaries $([0, x/2], [-y, y], [z_0, z_1])$ - green box) and the far sub-scene (with boundaries $([x/2, x], [-y, y], [z_0, z_1])$ - blue box). Red boxes correspond to ground-truth objects.}
    \label{fig:near-far-example}
\end{figure}

As it was mentioned before, voxel-based methods process the part of the scene inside the field-of-view of a camera placed close to the Lidar sensor, as shown in Figure \ref{fig:near-far-example}. We refer to the part of the scene processed by the voxel-based methods as \textit{full-scene} hereafter. Considering the fact that the features of each pillar are generated only based on the points inside it in an independent to the rest of the pillars manner and that the density of points inside pillars placed at different distances from the Lidar sensor in the scene is different, it is natural to split the full-scene in sub-scenes based on depth zones, i.e. to divide the full-scene with respect to the forward-axis, as illustrated in Figure \ref{fig:near-far-example} where two depth zones with the same dimensions are defined. 
Even though the \textit{near sub-scene} and the \textit{far sub-scene} have the same size in the 3D scene, the number of points belonging to each of them is very different due to the difference in distances between objects inside these two sub-scenes and the Lidar sensor. For instance, the near sub-scene on KITTI evaluation set used in TANet \cite{liu2019tanet} for class \textit{Car} contains $17,026$ points on average, while the far sub-scene contains only $1,127$ points on average. This means that the point cloud corresponding to the far sub-scene is sparser by a factor of $10$ compared to the point cloud of the near sub-scene. Having such a small number of points in the far sub-scene rises questions related to the efficiency of using voxel-based methods with voxelization grid of a fixed size for all locations of the 3D scene, as well as the object class multimodality inherited by the different levels of sparsity at different distances from the Lidar sensor. By comparing the pseudo-image generated by voxel-based methods corresponding to the full-scene and the pseudo-images corresponding to the near and far sub-scenes, it can be seen that the two latter pseudo-images correspond to two (non-overlapping) parts of the first pseudo-image, each having half of its size. As the point cloud in the far sub-scene is much sparser, the corresponding pseudo-image contains a large number of empty pillars. That is, the model needs to learn different representations for objects belonging to the same class (despite the fact that they may have very similar appearance and orientation) due to high differences in point cloud sparsity.

\begin{figure*}[!t]
    \includegraphics[width=1\linewidth]{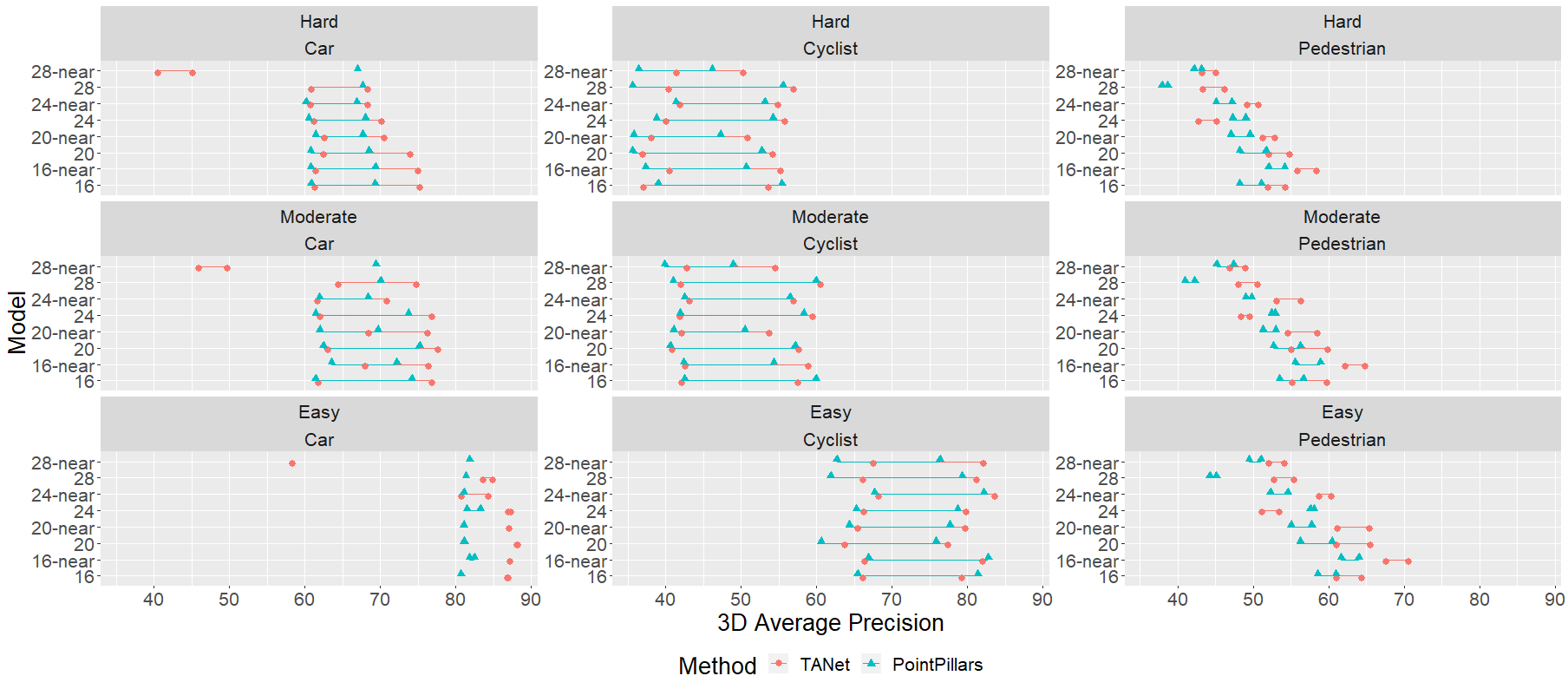}
    \caption{Performance evaluation of PointPillars and TANet models on the full-scene and the near sub-scene. Models with name $\{16, 20, 24, 28\}$ have a corresponding voxel size. Models with suffix ``-near" are trained on a near sub-scene. Each model is evaluated on the full-scene and the near sub-scene, therefore having 2 points per line, where the leftmost point corresponds to the near sub-scene evaluation and the rightmost point corresponds to the full-scene evaluation. Red circles represent AP values of TANet models and turquoise triangles represent AP values of PointPillars models. Red/turquoise lines represent difference in AP of the full-scene and near sub-scene evaluations.}
    \label{fig:comparison}
\end{figure*}

To determine the effect of adopting different sizes of scenes and pseudo-images in the performance and speed of voxel-based methods, we conduct an extensive evaluation based on the following steps:
\begin{itemize}
    \item We train models with pillar sizes $v_x = v_y = d$, with $d$ taking values in the set $\{16, 20, 24, 28\}$. For each pillar size, class combination and method, we train a model on objects appearing in the full-scene and another model trained on objects appearing in the near sub-scene. We use \textit{Car} and \textit{Pedestrian+Cyclist} class combinations as in the original PointPillars and TANet.
    \item We evaluate each model on the full-scene and on the near sub-scene. Therefore, each model is evaluated on scenes of size equal to those used during its training process and on scenes with a different size compared to the scenes used during its training process.
    \item When evaluating the models using objects belonging to the near sub-scene, we measure their performance considering all ground-truth objects in the full scene and considering only the ground-truth objects inside the near sub-scene.
\end{itemize}

By applying these experiments we can compare performance of trained models when applied to the full-scene and to the near sub-scene, calculating the drop of performance between these two cases. This alone cannot give full information about the ability of the models to detect objects in the far sub-scene due to influence of the uneven objects' class and difficulty distributions between near and far sub-scenes, and thus the evaluation of the models on the near sub-scene considering only ground-truth objects inside it can be used to determine the performance loss caused by not detecting the objects inside the far sub-scene. 

\section{Experiments}\label{sec:experiments}
We analyze the performance of models obtained using two voxel-based methods, i.e. PointPillars \cite{lang2018pointpillars} and TANet \cite{liu2019tanet}. We follow the configurations of these two methods, i.e. we use a scene with limits $[0, 69.12]$ for forward-axis, $[-39.68, 39.68]$ for left-right axis and $[-3, 1]$ for vertical axis for class \textit{Car}; and a scene with limits $[0, 47.36]$ for forward-axis, $[-19.84, 19.84]$ for left-right axis and $[-2.5, 0.5]$ for vertical axis for classes \textit{Pedestrian} and \textit{Cyclist}. These limits were designed for the voxel size $16$ and are slightly adjusted for the other voxel sizes, so that the resulting pseudo-images have a width and a height dimensions that are multiples of 8, which is required by the structure of the methods' FCN modules. The near sub-scene for class \textit{Car} has limits $[0, 34.56]$ for forward-axis, $[-39.68, 39.68]$ for left-right axis and $[-3, 1]$ for vertical axis, while for the classes \textit{Pedestrian} and \textit{Cyclist} it has limits $[0, 47.36]$ for forward-axis, $[-19.84, 19.84]$ for left-right axis and $[-2.5, 0.5]$ for vertical axis.

We train each models for 160 epochs with batch size 2 and evaluate on the $3,769$ samples from the evaluation subset of KITTI \cite{liu2019tanet}. The Average Precision (AP) \cite{2010pascal} metric is used to evaluate detection accuracy on the three object difficulty levels defined in the dataset, namely easy, moderate and hard. The object difficulty level depends on the size of its 2D projection on the camera plane and its occlusion level \cite{2012kitti}. Each model is evaluated on a desktop GPU NVIDIA GeForce GTX 1080Ti and the embedded platforms NVIDIA Jetson Tx2 and NVIDIA Jetson AGX Xavier. We use the MAXN power mode for both Tx2 and Xavier for maximum performance. The results of evaluation are given in Figure \ref{fig:comparison}.
\begin{figure*}[!t]
    \includegraphics[width=1\linewidth]{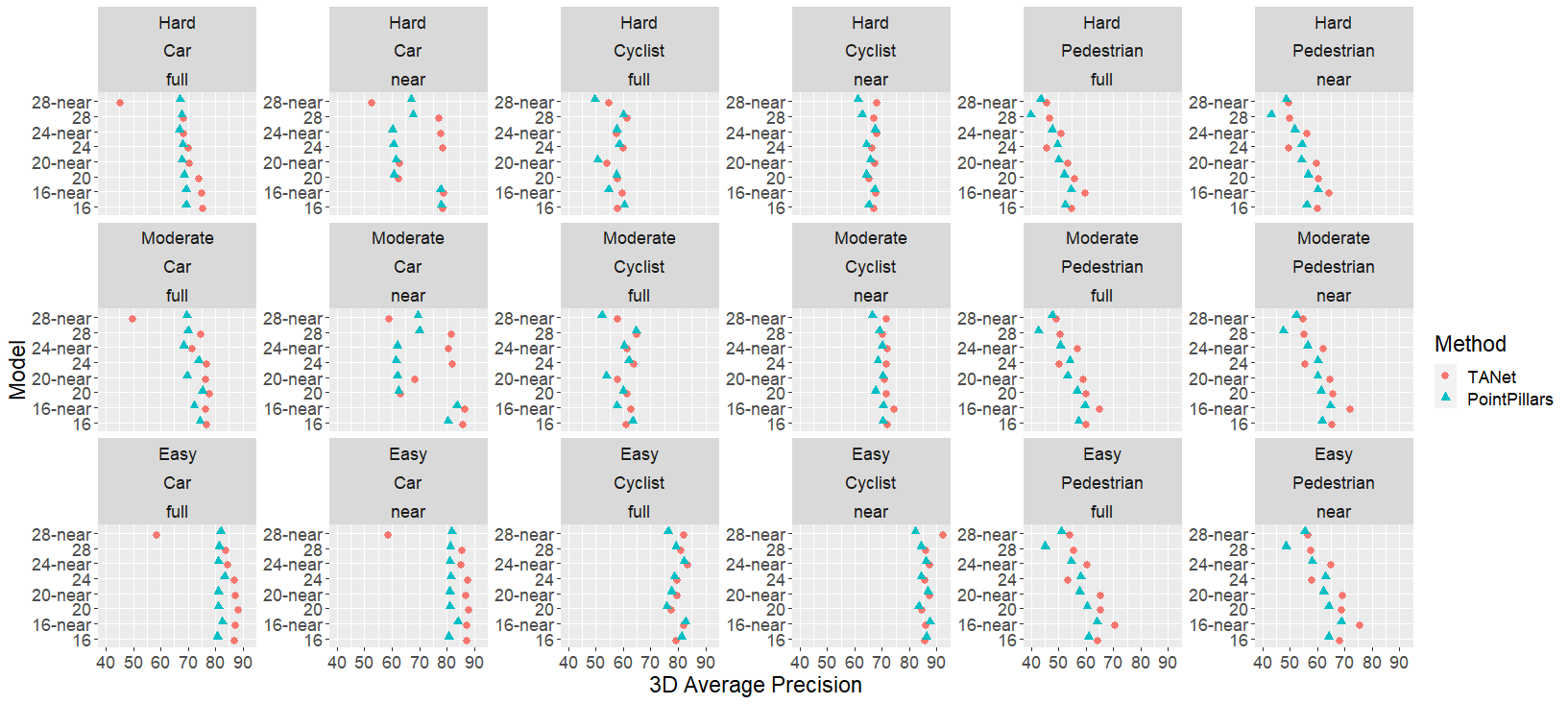}
    \caption{Performance Evaluation of PointPillars and TANet models on the full-scene and the near sub-scene considering only ground-truth objects inside the selected sub-scene. Models with names $\{16, 20, 24, 28\}$ have a corresponding voxel size and are trained on the full-scene. Models with suffix ``-near" are trained on the near sub-scene. Each model is evaluated on both the full-scene and the near sub-scene. Red circles represent AP values of TANet models and turquoise triangles represent AP values of PointPillars models.}
    \label{fig:comparison-1-1}
\end{figure*}

\begin{figure}[!ht]
    \includegraphics[width=1\linewidth]{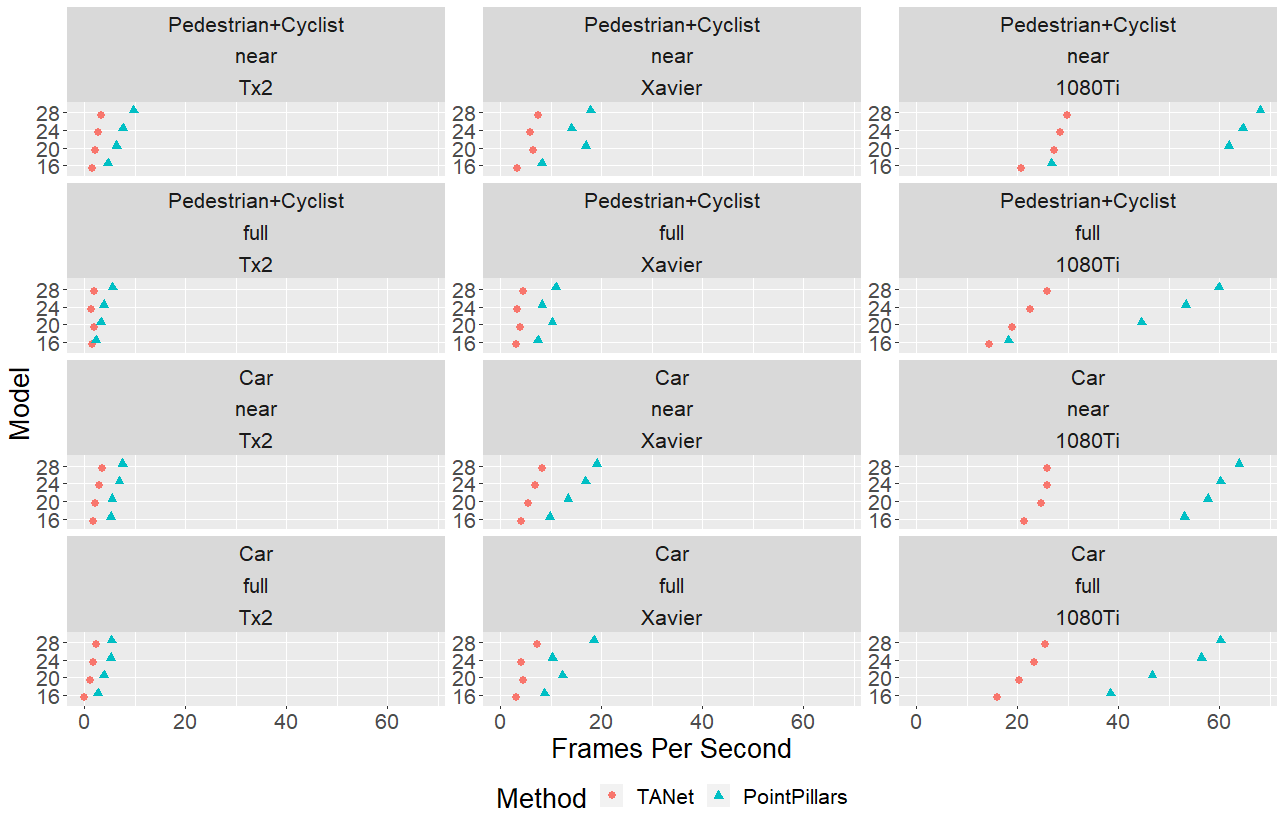}
    \caption{FPS evaluation of PointPillars and TANet models on the full-scene and the near sub-scene on desktop GPU and embedded systems. Models with names $\{16, 20, 24, 28\}$ have a corresponding voxel size. Red circles represent FPS values of TANet models and turquoise triangles represent FPS values of PointPillars models.}
    \label{fig:all-fps}
\end{figure}

\begin{table}[!ht]
    \caption{Distribution of object classes in the evaluation subset of KITTI dataset and object difficulties on a far sub-scene.}
    \label{tab:kitti-distributions}
    \centering
    \begin{tabular}{ccc}
        \toprule
        \textbf{Car} & \textbf{Pedestrian} & \textbf{Cyclist} \\
        \midrule
        75\% & 15\% & 4\% \\
        \midrule
        \multicolumn{3}{c}{\textbf{Far sub-scene}}\\
        \textbf{Easy} & \textbf{Moderate} & \textbf{Hard} \\
        14\% & 71\% & 15\% \\
        \bottomrule
    \end{tabular}
\end{table}

As can be seen in Figure \ref{fig:comparison}, the drop in performance when evaluating on the full-scene and near-scene is identical for each class across different difficulties, but is the lowest for objects with \textit{easy} difficulty level and class \textit{Car}. This can be explained by the distribution of class difficulties given in the Table \ref{tab:kitti-distributions}. Class \textit{Car} is the most represented in the dataset and difficulty level \textit{easy} is the least present in the far sub-scene, meaning that there is an insufficient number of selected objects to make a difference in performance.

Performance for class \textit{Pedestrian} is changing the least and this can be described either by the lack of objects belonging to this class in the far sub-scene, or the inability of models to detect these objects on the far sub-scene. Considering the difference between evaluations on the full-scene and the near sub-scene for class \textit{Pedestrian} from Figure \ref{fig:comparison-1-1}, we can conclude that there are enough objects belonging to class \textit{Pedestrian} on the far sub-scene to make a higher difference, but they remain undetected by the models.

Comparing the performance of models trained on the full-scene and on the near sub-scene, it can be seen that their results are almost identical but, in some cases, models trained on the near sub-scene lead to a better performance than the corresponding models trained on the scene (i.e. 16-near for \textit{Pedestrian+Cyclist}). However, these models have never been trained on objects of the far sub-scene, meaning that they apply features learned from near objects to far objects having a different point cloud structure. The fact that the model trained on near objects achieved better performance indicates that models trained on the full-scene fail to learn separate features for far objects and try to apply features learned by near objects on objects appearing in the full-scene.

As shown in Figure \ref{fig:all-fps}, application of a model on the near sub-scene leads to a $25\%$ increase in FPS on average on a desktop GPU, $40\%$ increase on an NVIDIA Jetson AGX Xavier and $60\%$ increase on an NVIDIA Jetson Tx2. Tx2 is the least powerful system among those tested, therefore only the $28$ PointPillars model applied on the near sub-scene can run in real-time for a Lidar with sampling rate of $10$ Hz. The $16$ PointPillars model when applied on the near sub-scene and $20$-$28$ PointPillars models applied on both full-scene and near sub-scene on Xavier can also be called real-time for a Lidar with sampling frequency of $10$ Hz. Near $28$ PointPillars model is close to real-time for a $20$ Hz Lidar.

\section{Conclusions and future work}\label{sec:conclusions}
In this paper, we analyzed the performance of two popular voxel-based 3D object detection methods providing a good compromise between high performance and speed based on their ability to detect objects appearing in locations of the scene that are far away from the Lidar sensor, and their speed when deployed on embedded platforms equipped with high-performance GPUs. Our analysis shows that these methods mostly fail to detect distant small objects due to the sparsity of the input point clouds at large distances. Moreover, models trained on near objects achieve similar or better performance compared to those trained on all objects in the scene. This means that the models learn object appearance representations mostly from near objects. Our findings suggest that a considerable part of the computations of existing methods is focused on locations of the scene that do not contribute with successful detection. This means that the methods can achieve a speed-up of $40$-$60\%$ by restricting operation to near objects while not sacrificing much in performance. A possible remedy towards addressing these limitations of voxel-based 3D object detection methods could be the application of complementary models that can achieve high performance on lower-resolution pseudo-images encoding the contents of the far sub-scene, in combination with the model operating on the near sub-scene, as this approach can lead to an increase in performance and in the total inference speed.

\section*{Acknowledgement}
This work has received funding from the European Union’s Horizon 2020 research and innovation programme under grant agreement No 871449 (OpenDR). This publication reflects the authors’ views only. The European Commission is not responsible for any use that may be made of the information it contains.

\appendix
\noindent
\underline{TANet inference speed:} During the evaluation of TANet\footnote{https://github.com/happinesslz/TANet} we noticed that the ratio of the inference speed between TANet and PointPillars were lower, than stated 30 to 60 FPS. TANet code has timers to count inference time of separate modules, counting total inference time for the evaluation pass and the number of samples processed. The average inference time is computed by dividing total time over the number of processed samples. We implemented additional timers that count FPS for each inference and the average FPS for the whole evaluation pass independently. The resulting FPS were quite different and the reason is that TANet counts each processed frame twice: once in $voxelnet.py::predict\_coarse$ and second time in $voxelnet.py::predict\_refine$, increasing the value $\_total\_inference\_count$ by $batch\_size$ every pass. Both functions are called to create final prediction, so the inference time is computed once per prediction, but the frames' counter is increased twice, reporting higher FPS than it should be.

\bibliographystyle{IEEEtran}
\bibliography{bibliography}

\begin{thebibliography}{10}
\providecommand{\url}[1]{#1}
\csname url@samestyle\endcsname
\providecommand{\newblock}{\relax}
\providecommand{\bibinfo}[2]{#2}
\providecommand{\BIBentrySTDinterwordspacing}{\spaceskip=0pt\relax}
\providecommand{\BIBentryALTinterwordstretchfactor}{4}
\providecommand{\BIBentryALTinterwordspacing}{\spaceskip=\fontdimen2\font plus
\BIBentryALTinterwordstretchfactor\fontdimen3\font minus
  \fontdimen4\font\relax}
\providecommand{\BIBforeignlanguage}[2]{{%
\expandafter\ifx\csname l@#1\endcsname\relax
\typeout{** WARNING: IEEEtran.bst: No hyphenation pattern has been}%
\typeout{** loaded for the language `#1'. Using the pattern for}%
\typeout{** the default language instead.}%
\else
\language=\csname l@#1\endcsname
\fi
#2}}
\providecommand{\BIBdecl}{\relax}
\BIBdecl

\bibitem{2018mf3d}
B.~Xu and Z.~Chen, ``{Multi-Level Fusion Based 3D Object Detection From
  Monocular Images},'' in \emph{{IEEE} Conference on Computer Vision and
  Pattern Recognition}, 2018.

\bibitem{2018monogrnet}
Z.~Qin, J.~Wang, and Y.~Lu, ``{MonoGRNet: {A} Geometric Reasoning Network for
  Monocular 3D Object Localization},'' in \emph{{AAAI} Conference on Artificial
  Intelligence}, 2019.

\bibitem{2019mono3dobjdet}
I.~{Barabanau}, A.~{Artemov}, E.~{Burnaev}, and V.~{Murashkin}, ``{Monocular 3D
  Object Detection via Geometric Reasoning on Keypoints},''
  \emph{arXiv:1905.05618}, 2019.

\bibitem{2019monodis}
A.~Simonelli, S.~R. Bul{\`{o}}, L.~Porzi, M.~Lopez{-}Antequera, and
  P.~Kontschieder, ``{Disentangling Monocular 3D Object Detection},'' in
  \emph{{IEEE/CVF} International Conference on Computer Vision}, 2019.

\bibitem{2018pseudolidar}
Y.~Wang, W.~Chao, D.~Garg, B.~Hariharan, M.~E. Campbell, and K.~Q. Weinberger,
  ``{Pseudo-LiDAR From Visual Depth Estimation: Bridging the Gap in 3D Object
  Detection for Autonomous Driving},'' in \emph{{IEEE} Conference on Computer
  Vision and Pattern Recognition}, 2019.

\bibitem{2019pseudolidar}
Y.~{You}, Y.~{Wang}, W.-L. {Chao}, D.~{Garg}, G.~{Pleiss}, B.~{Hariharan},
  M.~{Campbell}, and K.~Q. {Weinberger}, ``{Pseudo-LiDAR++: Accurate Depth for
  3D Object Detection in Autonomous Driving},'' in \emph{International
  Conference on Learning Representations}, 2020.

\bibitem{2020cdn}
C.~Li, J.~Ku, and S.~L. Waslander, ``{Confidence Guided Stereo 3D Object
  Detection with Split Depth Estimation},'' \emph{arXiv:2003.05505}, 2020.

\bibitem{2017voxelnet}
Y.~Zhou and O.~Tuzel, ``{VoxelNet: End-to-End Learning for Point Cloud Based 3D
  Object Detection},'' in \emph{{IEEE} Conference on Computer Vision and
  Pattern Recognition}, 2018.

\bibitem{lang2018pointpillars}
A.~H. Lang, S.~Vora, H.~Caesar, L.~Zhou, J.~Yang, and O.~Beijbom,
  ``{PointPillars: Fast Encoders for Object Detection from Point Clouds},'' in
  \emph{{IEEE} Conference on Computer Vision and Pattern Recognition}, 2019.

\bibitem{liu2019tanet}
Z.~Liu, X.~Zhao, T.~Huang, R.~Hu, Y.~Zhou, and X.~Bai, ``{TANet: Robust 3D
  Object Detection from Point Clouds with Triple Attention},'' in \emph{{AAAI}
  Conference on Artificial Intelligence}, 2020.

\bibitem{2019hotspotnet}
Q.~Chen, L.~Sun, Z.~Wang, K.~Jia, and A.~L. Yuille, ``{Object as Hotspots: An
  Anchor-Free 3D Object Detection Approach via Firing of Hotspots},'' in
  \emph{European Conference on Computer Vision}, 2020.

\bibitem{2017frustumpointnet}
C.~R. Qi, W.~Liu, C.~Wu, H.~Su, and L.~J. Guibas, ``{Frustum PointNets for 3D
  Object Detection From {RGB-D} Data},'' in \emph{2018 {IEEE} Conference on
  Computer Vision and Pattern Recognition}, 2018.

\bibitem{2019frustumconvnet}
Z.~Wang and K.~Jia, ``{Frustum ConvNet: Sliding Frustums to Aggregate Local
  Point-Wise Features for Amodal},'' in \emph{{IEEE/RSJ} International
  Conference on Intelligent Robots and Systems}, 2019.

\bibitem{2019pointpainting}
S.~Vora, A.~H. Lang, B.~Helou, and O.~Beijbom, ``{PointPainting: Sequential
  Fusion for 3D Object Detection },'' in \emph{{IEEE/CVF} Conference on
  Computer Vision and Pattern Recognition}, 2020.

\bibitem{2020gslamtc}
X.~He, S.~Bai, J.~Chu, and X.~Bai, ``{An Improved Multi-View Convolutional
  Neural Network for 3D Object Retrieval},'' \emph{{IEEE} Transactions on Image
  Processing}, vol.~29, pp. 7917--7930, 2020.

\bibitem{2015mvcnn}
H.~Su, S.~Maji, E.~Kalogerakis, and E.~G. Learned{-}Miller, ``{Multi-view
  Convolutional Neural Networks for 3D Shape Recognition},'' in \emph{{IEEE}
  International Conference on Computer Vision}, 2015.

\bibitem{2017squeezeseg}
B.~Wu, A.~Wan, X.~Yue, and K.~Keutzer, ``{SqueezeSeg: Convolutional Neural Nets
  with Recurrent {CRF} for Real-Time Road-Object Segmentation from 3D LiDAR
  Point Cloud},'' in \emph{{IEEE} International Conference on Robotics and
  Automation}, 2018.

\bibitem{2016velofcn}
B.~Li, T.~Zhang, and T.~Xia, ``{Vehicle Detection from 3D Lidar Using Fully
  Convolutional Network},'' in \emph{Robotics: Science and Systems XII}, 2016.

\bibitem{2020rtclusterlidar}
M.~{Verucchi}, L.~{Bartoli}, F.~{Bagni}, F.~{Gatti}, P.~{Burgio}, and
  M.~{Bertogna}, ``{Real-Time clustering and LiDAR-camera fusion on embedded
  platforms for self-driving cars},'' in \emph{IEEE International Conference on
  Robotic Computing}, 2020.

\bibitem{rashed2019fuseMODNet}
H.~Rashed, M.~Ramzy, V.~Vaquero, A.~E. Sallab, G.~Sistu, and S.~Yogamani,
  ``{FuseMODNet: Real-Time Camera and LiDAR based Moving Object Detection for
  robust low-light Autonomous Driving},'' in \emph{International Conference on
  Computer Vision Workshops}, 2019.

\bibitem{2020vislidar}
M.~{Sualeh} and G.~W. {Kim}, ``{Visual-LiDAR Based 3D Object Detection and
  Tracking for Embedded Systems},'' \emph{IEEE Access}, vol.~8, pp.
  156\,285--156\,298, 2020.

\bibitem{2020mobilecenternet}
J.~{Yu}, H.~{Xie}, M.~{Li}, G.~{Xie}, Y.~{Yu}, and C.~W. {Chen}, ``{Mobile
  Centernet for Embedded Deep Learning Object Detection},'' in \emph{IEEE
  International Conference on Multimedia and Expo Workshops}, 2020.

\bibitem{2020efyolo}
Z.~{Wang}, J.~{Zhang}, Z.~{Zhao}, and F.~{Su}, ``{Efficient Yolo: A Lightweight
  Model For Embedded Deep Learning Object Detection},'' in \emph{IEEE
  International Conference on Multimedia and Expo Workshops}, 2020.

\bibitem{2020quadaprox}
X.~{Yang}, S.~{Chaudhuri}, L.~{Naviner}, and L.~{Likforman}, ``{Quad-Approx
  CNNs for Embedded Object Detection Systems},'' in \emph{IEEE International
  Conference on Electronics, Circuits and Systems}, 2020.

\bibitem{2020quanttiling}
G.~{Plastiras}, S.~{Siddiqui}, C.~{Kyrkou}, and T.~{Theocharides}, ``{Efficient
  Embedded Deep Neural-Network-based Object Detection Via Joint Quantization
  and Tiling},'' in \emph{IEEE International Conference on Artificial
  Intelligence Circuits and Systems}, 2020.

\bibitem{2012kitti}
A.~Geiger, P.~Lenz, and R.~Urtasun, ``{Are we ready for autonomous driving? The
  {KITTI} vision benchmark suite},'' in \emph{{IEEE} Conference on Computer
  Vision and Pattern Recognition}, 2012.

\bibitem{nuScenes2019}
H.~Caesar, V.~Bankiti, A.~H. Lang, S.~Vora, V.~E. Liong, Q.~Xu, A.~Krishnan,
  Y.~Pan, G.~Baldan, and O.~Beijbom, ``{nuScenes: {A} Multimodal Dataset for
  Autonomous Driving},'' in \emph{{IEEE/CVF} Conference on Computer Vision and
  Pattern Recognition}, 2020.

\bibitem{2014fcn}
E.~Shelhamer, J.~Long, and T.~Darrell, ``{Fully Convolutional Networks for
  Semantic Segmentation},'' \emph{{IEEE} Transactions on Pattern Analysis and
  Machine Intelligence}, vol.~39, no.~4, pp. 640--651, 2017.

\bibitem{2010pascal}
M.~Everingham, L.~Gool, C.~K. Williams, J.~Winn, and A.~Zisserman, ``{The
  Pascal Visual Object Classes (VOC) Challenge},'' \emph{International Journal
  of Computer Vision}, vol.~88, no.~2, pp. 303–--338, 2010.

\end{thebibliography}
\end{document}